\pdfoutput=1

\documentclass[11pt]{article}

\usepackage[final]{acl}

\usepackage{times}
\usepackage{latexsym}

\usepackage[T1]{fontenc}

\usepackage[utf8]{inputenc}

\usepackage{microtype}

\usepackage{inconsolata}

\usepackage{graphicx}
\usepackage{amsmath}
\usepackage{booktabs}
\usepackage{arydshln}
\usepackage{amsfonts}
\usepackage{dblfloatfix}
\usepackage{multirow}
\usepackage{tabularx}
\usepackage{xcolor}
\usepackage{float}
\usepackage{enumitem}
\usepackage{adjustbox}
\usepackage{needspace}
\usepackage[colorinlistoftodos]{todonotes}

\newcommand{\Llama}{\mbox{Llama-3.1-8B}}
\newcommand{\Qwen}{\mbox{Qwen3-8B}}
\newcommand{\Gemma}{\mbox{Gemma-3-4B-IT}}
\newcommand{\rgg}{robustness--generation}
\newcommand{\rqg}{reward--quality}
\newcommand{\SFTd}{SFT\textsubscript{d}}

\title{DiaLLM: An Investigation into the \\ Robustness--Generation Gap in English Dialect Adaptation}

\author{
  Jordan Painter$^1$ \quad Dipankar Srirag$^2$ \quad Adarsh Kappiyath$^1$ \\
  \textbf{Diptesh Kanojia$^1$ \quad Aditya Joshi$^2$ \quad Lu Yin$^1$} \\[4pt]
  $^1$Institute for People-Centred AI, University of Surrey, Surrey, United Kingdom \\
  $^2$University of New South Wales, Sydney, Australia \\[4pt]
  \texttt{\{j.painter,a.kappiyath,d.kanojia,l.yin\}@surrey.ac.uk} \quad \\
  \texttt{\{d.srirag,aditya.joshi\}@unsw.edu.au}
}

\begin{document}
\maketitle
\begin{abstract}
Large language models increasingly \emph{understand} dialectal English, yet still \emph{produce} only standard, US-leaning English, leaving dialectal generation, the harder half of the problem, largely unaddressed. We introduce \textbf{DiaLLM}, which continually pretrains three open-weight language model families on the International Corpus of English and applies implicit and explicit post-training paradigms, each combined with three model alignment strategies, giving the first controlled comparison of these components across Australian, Indian, and Northern British English. Our results reveal a \emph{robustness–generation gap}: benchmarks are shaped by continual pretraining and SFT, while alignment visibly reshapes generation in ways benchmarks do not capture. Explicit variety-targeted adaptation produces output reliably recognised as dialectal and judged more dialectal than broad alignment, yet where human judgement was directly assessed, the method that most aggressively optimises the dialectal reward is not the one judged most dialectal. Independent linguistic analysis corroborates this \rqg{} gap, most clearly on two of the three families. No single alignment method dominates, and closing the gap will require richer reward designs and continued investment in dialectal resources. We release all code, checkpoints, and preference datasets.
\end{abstract}

\section{Introduction}
\label{sec:intro}
\begin{figure}[t!]
  \centering
  \includegraphics[width=0.9\columnwidth]{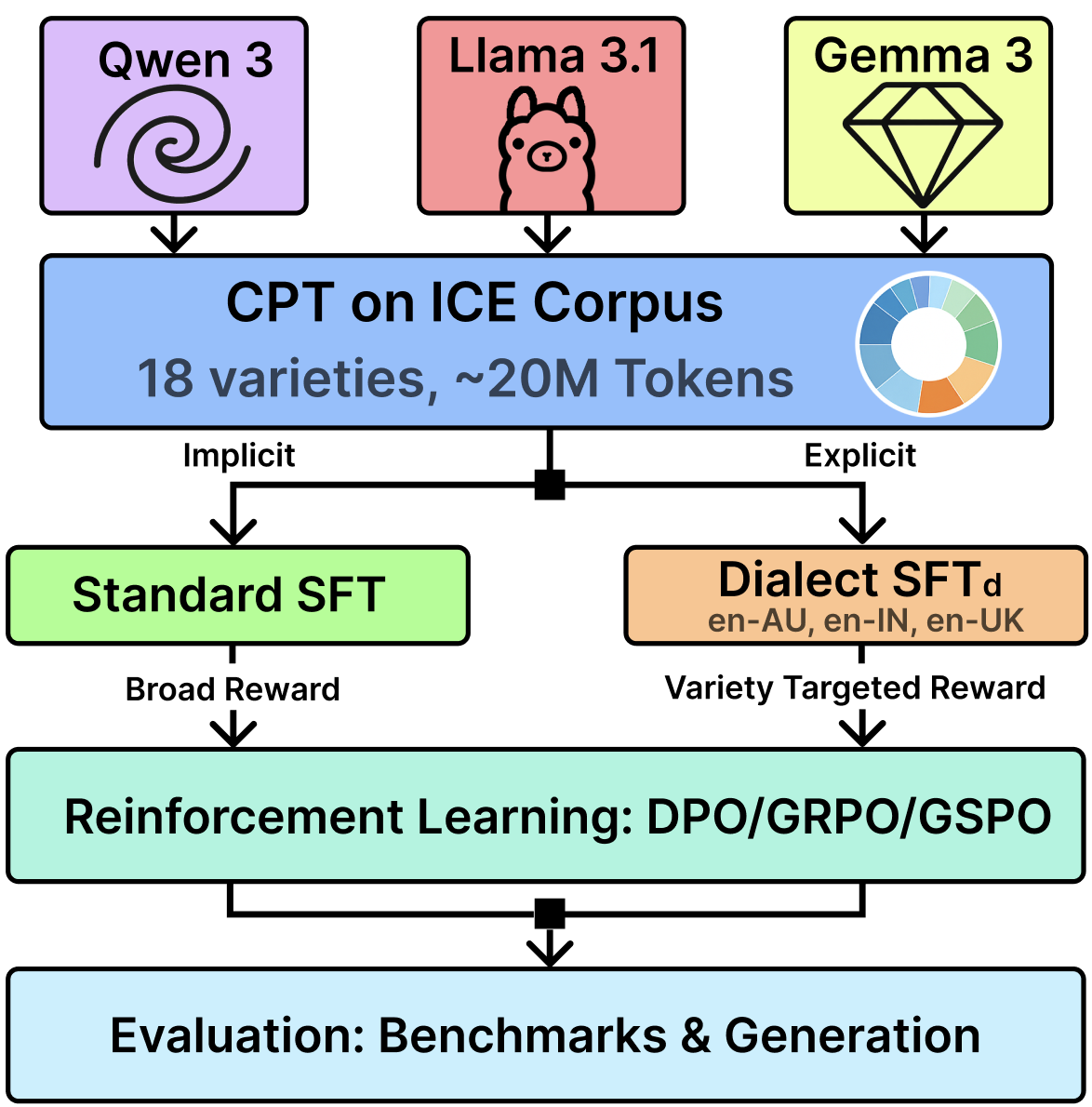}
  \caption{Overview of the \textbf{DiaLLM} pipeline: continual pretraining on ICE followed by either \textit{implicit adaptation} (standard SFT + alignment) or \textit{explicit adaptation} (dialectal SFT + variety-targeted alignment), with DPO, GRPO, and GSPO compared across both paradigms.}
  \label{fig:pipeline}
\end{figure}

Large language models (LLMs) achieve strong performance across a range of NLP tasks \cite{wang2024mmluprorobustchallengingmultitask,touvron2023llamaopenefficientfoundation}, but often falter on dialectal and non-standard varieties that diverge from standardised norms \cite{joshi2025natural,blodgett-etal-2016-demographic,hofmann2024dialectprejudicepredictsai,mire-etal-2025-rejected}. These disparities reflect structural imbalances in training data: large-scale corpora privilege majority language forms, marginalising non-standard varieties \cite{10.1145/3442188.3445922,joshi-etal-2020-state}. While domain adaptation has received substantial attention \cite{gururangan2020dontstoppretrainingadapt}, robustness to dialectal variation within a single language remains comparatively underexplored. Existing work is concentrated heavily on AAVE and code-switching \citep{blodgett-etal-2016-demographic,sap-etal-2019-risk,hofmann2024dialectprejudicepredictsai}, while regional varieties such as Australian, Indian, and Northern British English receive substantially less attention at the alignment stage. Methods that do address dialectal variation, such as TADA \citep{held-etal-2023-tada} and HyperLoRA \citep{xiao-etal-2023-task}, do so via dialect adapters or low-rank components targeting NLU robustness under dialectal \textit{input}. LoRDD \citep{srirag-etal-2025-predicting} similarly employs a low-rank dialect adapter for decoder models in a task-specific setting. None of them investigates \textit{full-pipeline adaptation} across pretraining, fine-tuning, and alignment, nor whether models can produce dialectally appropriate \textit{output}.

Dialectal generation (producing a language variety on output, not merely tolerating it on input) is a critical part of the problem that current methods do not address. Consider three users chasing a late parcel:
\begin{quote}
    \textit{Indian English:} My parcel is not \textbf{yet coming}; 
    kindly \textbf{do the needful} and \textbf{prepone} the delivery.\\[4pt]
    \textit{Northern British English:} \textbf{Me} parcel's not arrived; 
    \textbf{summat} were up \textbf{wi'} the courier.\\[4pt]
    \textit{Australian English:} Me parcel still hasn't turned up, 
    \textbf{reckon} it's good this \textbf{arvo, eh?}
\end{quote}
\noindent A modern LLM will \emph{understand} all three, yet answer each in flattened Standard American English \citep{fleisig-etal-2024-linguistic}. A model that understands every variety but speaks only one has not solved the dialect problem; it has hidden half of it. This gap matters not because every speaker necessarily requires a response in their own variety, but because they should have the option.

We introduce \textbf{DiaLLM}, a dialect adaptation framework that continually pretrains three open-weight LLM families (\Llama{}, \Qwen{}, \Gemma{}) on the International Corpus of English (ICE; \citealt{greenbaum1996ice}), spanning 18 English varieties. As shown in Figure~\ref{fig:pipeline}, from each CPT checkpoint we pursue two post-training paradigms: \textit{implicit adaptation} applies standard instruction-tuning and broad alignment; \textit{explicit adaptation} uses dialectally-perturbed supervision and variety-targeted alignment. Across both paradigms, we compare Direct Preference Optimisation (DPO), Group Relative Policy Optimisation (GRPO), and Group Sequence Policy Optimisation (GSPO). We treat the relative effectiveness of each pipeline component and alignment method as an empirical question rather than assuming any single recipe is best.

Our results reveal a \textit{\rgg{}} gap: benchmark performance is driven primarily by CPT and SFT, while alignment visibly shapes generation in ways benchmarks do not capture. Explicit variety-targeted adaptation is reliably recognised as dialectal and judged more dialectal than broad alignment by both human annotators and LLM judges. Yet, the method that most aggressively optimises the dialectal reward is not the one judged most dialectal. Our linguistic analysis confirms that this method produces the fewest recognisable dialectal markers across all three families, showing a mismatch between feature-density reward and perceived dialectal quality.

We make three primary contributions:
\begin{enumerate}
\item \textbf{DiaLLM}, a full-pipeline dialect adaptation framework with a controlled comparison of continual pretraining, supervised fine-tuning, and three alignment methods (DPO, GRPO, GSPO) across two post-training paradigms and three open-weight LLM families. We release all code (including a linguistic-analysis toolkit), model checkpoints, and dialectal preference datasets\footnote{\url{https://github.com/jordanpainter/diallm}}. 
\item We establish a \textbf{\rgg{} gap}: benchmark performance is set by CPT and SFT, whereas alignment moves benchmarks only marginally, yet visibly reshapes generation, showing benchmarks alone misjudge what alignment does.
\item We expose a \textbf{\rqg{} gap}: although explicit, variety-targeted adaptation is reliably recognised as dialectal and judged more dialectal than broad alignment, the method that most aggressively optimises the dialectal reward (GRPO) is \emph{not} the one judged most dialectal by LLM judges or human annotators on \Llama{}, where this evaluation is conducted; an independent linguistic analysis across all three families corroborates this finding.
\end{enumerate}

\section{Related Work}
\label{sec:related}
LLMs systematically underperform on dialectal and non-standard English: standard NLP tools fail on AAVE \citep{blodgett-etal-2016-demographic}, dialectal text is disproportionately flagged as toxic \citep{sap-etal-2019-risk}, and LLMs encode racio-linguistic stereotypes and generate lower-quality outputs for dialectal prompts even after alignment \citep{hofmann2024dialectprejudicepredictsai,mire-etal-2025-rejected}, with reward models themselves steering outputs toward White Mainstream English \citep{mire-etal-2025-rejected}. These gaps are documented at scale across 281 varieties using VALUE \citep{ziems2022valueunderstandingdialectdisparity}, BESSTIE \citep{srirag2025besstiebenchmarksentimentsarcasm}, and DialectBench \citep{faisal2024dialectbenchnlpbenchmarkdialects}. \citet{maheshwari-etal-2026-improving} similarly target robustness via LoRA and mixture-of-experts adapters, but for classification rather than generation.

Methods that do address dialectal variation---TADA \citep{held-etal-2023-tada} and HyperLoRA \citep{xiao-etal-2023-task}---do so via dialect adapters or low-rank components targeting NLU robustness under dialectal \textit{input}; LoRDD \citep{srirag-etal-2025-predicting} similarly employs a low-rank dialect adapter for decoder models in a task-specific setting. None of these existing works investigates full-pipeline adaptation across pretraining, fine-tuning, and alignment paradigms, nor whether models can produce dialectally appropriate \textit{output}.

Continual pretraining on in-domain text is a well-established adaptation foundation \citep{gururangan2020dontstoppretrainingadapt}, applied across medical, legal, and multilingual settings \citep{xie2024mellama,huang2023lawyerllamatechnicalreport}; \citet{khan2025frenchdialectcpt} show dialect gaps in Québec French can be narrowed via CPT with limited benchmark degradation. For post-training, DPO \citep{rafailov2024directpreferenceoptimizationlanguage}, GRPO \citep{shao2024deepseekmathpushinglimitsmathematical}, and GSPO \citep{zheng2025groupsequencepolicyoptimization} offer contrasting approaches to preference optimisation; \citet{delangis2024multistyle} demonstrate that RL with automatic linguistic reward signals can steer generation toward stylistically defined targets---a principle we extend to dialect-specific features. None of these methods have been systematically compared for dialect adaptation.

\section{Methodology}
\label{sec:method}
\subsection{Base Models}

We apply our dialect adaptation pipeline to three open-weight LLM families spanning the 4B--8B parameter range: \textbf{\Llama{}}~\citep{grattafiori2024llama3herdmodels}, \textbf{\Qwen{}}~\citep{yang2025qwen3technicalreport}, and \textbf{\Gemma{}}~\citep{gemmateam2025gemma3technicalreport}\footnote{For \Gemma{}, we initialise CPT from the instruction-tuned variant rather than the base model, following poor performance in preliminary experiments with the base checkpoint; the remaining pipeline is identical.}. Selecting architecturally diverse models allows us to assess whether gains from dialect adaptation are robust across training regimes and vocabulary choices, rather than specific to a single architecture.

\subsection{Continual Pretraining}


\textbf{Corpus}: We continually pretrain all models using the International Corpus of English (ICE), a collection of comparable corpora from over 25 English-speaking regions. Each variety contains one million words of spoken and written language sampled using consistent sociolinguistic criteria, making ICE uniquely suited for this task: it is the only available multi-variety corpus spanning both inner and outer circle varieties at a usable scale, with no comparable resource covering this range of regional Englishes.

For this work, we include 18 ICE varieties with cleaned, digitised data available.\footnote{See Appendix~\ref{appendix:corpdeets} for preprocessing and filtering details. We could only obtain permissions to use the corpus for this set.} Access to further varieties was sought but could not be obtained, either because the relevant ICE teams did not respond to requests or because the data was not in a usable digitised form. The corpus covers a broad range of regional, cultural, and postcolonial Englishes, including Hong Kong, Nigeria, India, Jamaica, and others. In total, our training corpus comprises approximately 20 million tokens.

\textbf{Pretraining setup}: We continually pretrain each selected model on the cleaned ICE corpus before any 
supervised or preference-based post-training using GaLore \citep{pmlr-v235-zhao24s}, enabling memory-efficient full-parameter optimisation.
CPT is implemented as causal language modelling over the cleaned ICE text; full training details are provided in Appendix~\ref{appendix:hyperparams}.

\subsection{Post-Training Dataset}
\label{sec:posttraindata}

For post-training, we use the preference dataset introduced by \citet{cui2023ultrafeedback}, which contains high-quality prompt-response pairs annotated with binary preferences over completions. For the dialect-specific training thread, we convert the preferred completions into dialectal variants using the Multi-VALUE library \citep{ziems-etal-2023-multi}, targeting three English varieties: Australian English, Indian English, and Northern British English. Multi-VALUE applies transformations based on eWAVE features \citep{ewave}, which constitute the only feature-level inventory of morphosyntactic variation spanning all English varieties considered in this work. This makes eWAVE a natural foundation for both dialect perturbation and reward specification. During conversion, code blocks are automatically detected and preserved verbatim. The resulting datasets contain approximately 11.8k, 15.4k, and 18.4k preference pairs for Australian, Northern British, and Indian English, respectively. Differences in dataset size arise from unsuccessful conversions produced by Multi-VALUE.

\subsection{Alignment Strategy}
\label{sec:alignment}

From each CPT checkpoint, we pursue two dialect adaptation paradigms: the \textbf{implicit thread}  (broad training, with no variety targeting) and the \textbf{explicit thread} (single-variety targeting throughout), as illustrated in Figure~\ref{fig:pipeline}. We treat each stage as serving a distinct role in the pipeline: continual pretraining introduces dialectal exposure that is otherwise absent from models trained predominantly on standard English corpora; post-training methods can shape output distributions but cannot introduce dialectal signal that is not already latent in the model's weights; supervised fine-tuning restores general instruction-following behaviour.

\subsubsection{Supervised Fine-Tuning}

Both adaptation threads apply SFT from the CPT checkpoint, training only on chosen completions in a single-turn instruction format with early stopping based on held-out evaluation loss. Full hyperparameters are reported in Appendix~\ref{appendix:hyperparams}.

The \textbf{implicit thread} fine-tunes on the standard \textit{ultrafeedback-binarized-preferences} corpus, providing a general instruction-following objective with no explicit dialectal signal. The \textbf{explicit thread} fine-tunes instead on the dialect-specific dataset for the target variety, so the chosen completions already contain Multi-VALUE dialectal features. No general SFT stage intervenes between CPT and dialect SFT in the explicit thread, preserving the dialectal signal introduced by CPT.

\subsubsection{Direct Preference Optimisation}
\label{sec:dpo}

DPO \citep{rafailov2024directpreferenceoptimizationlanguage} is an offline method that optimises directly from contrastive pairs without querying a reward model. Preference pairs are drawn from the dialectal datasets in Section~\ref{sec:posttraindata}: chosen completions are Multi-VALUE dialectal variants; rejected completions are the original standard English forms. In the implicit thread, DPO trains on the combined three-variety dataset; the three varieties are pooled rather than targeted individually, preserving the variety-agnostic objective of this thread. In the explicit thread, only the target-variety subset is used.
\subsubsection{Dialect Feature Classifier}
\label{sec:classifier}

To provide a reward signal for online alignment, we train a multi-label classifier to detect 135 morphosyntactic and lexical features of dialectal English derived from the eWAVE typological database \citep{kortmann_electronic_2013}. eWAVE provides the only structured feature-level typology of morphosyntactic variation spanning our three target varieties, making it the sole available basis for a linguistically interpretable reward signal at this level of dialectal specificity. Full architecture, training, and calibration details are given in Appendix~\ref{appendix:classifier}. The classifier’s logits are passed through a sigmoid and the resulting probabilities are aggregated as $\phi_{\text{dial}}(y) = \log(1 + \sum_{i \in \mathcal{F}} \sigma(\text{logit}_i))$, where $\mathcal{F}$ is the active feature set: all 135 features in the implicit thread, or the attested variety-specific subset in the explicit thread (en-AU: 20 features; en-UK: 30 features; en-IN: 38 features).

\subsubsection{Reward Formulation}
\label{sec:reward}

Both GRPO and GSPO optimise a shared reward combining dialectal feature density and meaning preservation:
\begin{equation}
R = \lambda \cdot \phi_{\text{dial}}(y)
+ \tfrac{1-\lambda}{2} \cdot \phi_{\text{comet}}(y, \hat{y})
+ \tfrac{1-\lambda}{2} \cdot \phi_{\text{cos}}(y, \hat{y})
\end{equation}
$\phi_{\text{dial}}(y)$ is the log-sum dialectal reward from the feature classifier (Section~\ref{sec:classifier}). $\phi_{\text{comet}}(y, \hat{y})$ and $\phi_{\text{cos}}(y, \hat{y})$ measure semantic fidelity against the SFT chosen completion $\hat{y}$ using COMET \citep{rei-etal-2022-comet} and cosine similarity over \texttt{all-MiniLM-L6-v2} \citep{reimers2019sentencebertsentenceembeddingsusing} respectively. We fixed $\lambda = 0.80$ across all model families and alignment methods; 
full selection details are given in Appendix~\ref{appendix:hyperparams}. All components are z-score normalised.

\subsubsection{GRPO and GSPO}
\label{sec:grpo-gspo}

GRPO \citep{shao2024deepseekmathpushinglimitsmathematical} generates a group of rollouts per prompt, estimates a baseline from the group mean reward, and computes token-level policy gradient updates without a value network. GSPO \citep{zheng2025groupsequencepolicyoptimization} replaces token-level importance sampling with sequence-level reweighting, assigning credit by the full-sequence probability ratio under the current and reference policies, avoiding the bias introduced by per-token clipping. Both use four completions per prompt and the reward in Section~\ref{sec:reward}.

\section{Experiment Setup}
\label{sec:setup}

\subsection{General Benchmarks}

We evaluate general capability using BBH \citep{suzgun2022challengingbigbenchtaskschainofthought}, GPQA \citep{rein2023gpqagraduatelevelgoogleproofqa}, and GLUE \citep{wang-etal-2018-glue}, implemented via the \textit{EleutherAI/lm-evaluation-harness} \citep{eval-harness} with unmodified task templates.

\begin{table*}[ht]
\begin{adjustbox}{width=\linewidth, center}
\begin{tabular}{lcccccccccccc}
\toprule
& \multicolumn{4}{c}{\textbf{\Qwen{}}} & \multicolumn{4}{c}{\textbf{\Llama{}}} & \multicolumn{4}{c}{\textbf{\Gemma{}}} \\
\cmidrule(lr){2-5} \cmidrule(lr){6-9} \cmidrule(lr){10-13}
\textbf{Benchmark} & \SFTd{} & DPO & GRPO & GSPO & \SFTd{} & DPO & GRPO & GSPO & \SFTd{} & DPO & GRPO & GSPO \\
\midrule
\multicolumn{13}{l}{\textit{en-AU (Australian English)}} \\
GLUE     & 61.59 & 61.08 & 61.60 & \textbf{61.65} & 46.26 & 47.11 & 47.08 & \textbf{47.31} & 52.69 & \textbf{53.39} & 51.98 & 52.95 \\
VALUE    & \textbf{55.94} & 50.00 & 55.04 & 55.59 & 45.42 & \textbf{47.06} & 45.75 & 45.61 & 50.59 & 47.41 & 49.28 & \textbf{50.60} \\
BST-Sent & \textbf{90.02} & 89.57 & 89.93 & 89.92 & 87.31 & 87.26 & \textbf{87.35} & 87.21 & \textbf{84.91} & 84.73 & \textbf{84.91} & 84.76 \\
BST-Sarc & 37.07 & \textbf{37.36} & 36.84 & 36.52 & 39.45 & \textbf{41.18} & 40.54 & 38.95 & 27.31 & \textbf{31.86} & 28.90 & 28.30 \\
DID      & \textbf{67.11} & 65.44 & 66.61 & 64.94 & 67.95 & 64.44 & 67.11 & \textbf{69.28} & 35.56 & \textbf{35.73} & 35.56 & 35.56 \\
\midrule
\multicolumn{13}{l}{\textit{en-IN (Indian English)}} \\
GLUE     & 62.71 & \textbf{63.03} & 62.50 & 62.84 & 54.27 & \textbf{56.45} & 53.94 & 53.88 & 57.15 & 57.63 & 57.12 & \textbf{58.38} \\
VALUE    & 52.92 & 50.41 & \textbf{52.94} & 52.90 & 48.32 & \textbf{49.34} & 47.88 & 48.01 & 50.21 & 49.49 & 50.68 & \textbf{50.78} \\
BST-Sent & 89.75 & 89.61 & 89.78 & \textbf{90.01} & 81.52 & 80.95 & \textbf{82.64} & 82.09 & 77.50 & 75.53 & 78.97 & \textbf{79.11} \\
BST-Sarc & 39.27 & 38.50 & \textbf{39.43} & 38.87 & 36.36 & \textbf{37.68} & 35.47 & 35.27 & 32.79 & 32.54 & 32.88 & \textbf{33.22} \\
DID      & \textbf{71.45} & 70.12 & 70.79 & 70.45 & 64.78 & \textbf{65.78} & 64.11 & 63.44 & 13.02 & 13.02 & 13.02 & \textbf{13.36} \\
\midrule
\multicolumn{13}{l}{\textit{en-UK (Northern British English)}} \\
GLUE     & 63.48 & 63.42 & \textbf{63.99} & 63.72 & \textbf{54.05} & 53.77 & 53.95 & 53.78 & 56.03 & 55.35 & 55.45 & \textbf{56.42} \\
VALUE    & 58.27 & 54.59 & \textbf{58.50} & 58.49 & 45.67 & \textbf{48.35} & 45.56 & 45.59 & \textbf{48.93} & 46.36 & 48.37 & 48.61 \\
BST-Sent & 87.92 & \textbf{88.01} & \textbf{88.01} & 87.99 & 79.45 & \textbf{80.83} & 80.11 & 79.89 & 82.53 & 79.31 & 82.23 & \textbf{83.49} \\
BST-Sarc & 35.44 & 34.97 & 35.28 & \textbf{35.48} & 34.96 & 34.58 & \textbf{35.16} & 34.73 & \textbf{33.95} & 31.68 & 33.11 & 32.61 \\
DID      & 63.94 & 63.44 & 64.27 & \textbf{64.61} & 50.58 & \textbf{58.60} & 49.58 & 51.59 & 35.23 & \textbf{35.39} & 35.23 & 35.23 \\
\bottomrule
\end{tabular}
\end{adjustbox}
\caption{Explicit thread alignment comparison (\%). \SFTd{} denotes dialect-specific supervised fine-tuning; DPO, GRPO, and GSPO are applied to \SFTd{} with variety-targeted rewards. Bold indicates the best result within each model family per row and dialect group. BST-Sent and BST-Sarc refer to BESSTIE sentiment and sarcasm respectively.}
\label{tab:explicit-alignment}
\end{table*}

\subsection{Dialectal Benchmarks}

We evaluate dialectal robustness using three benchmark suites. The dialect
identification (DID) task from \textsc{DialectBench}
\citep{faisal2024dialectbenchnlpbenchmarkdialects} requires classifying
inputs as British English (en-GB), American English (en-US), or generic
English (en), with higher scores indicating greater sensitivity to dialectal variation; as DID contains no en-AU or en-IN class, scores for these conditions index general dialectal sensitivity rather than variety-specific recognition. VALUE \citep{ziems2022valueunderstandingdialectdisparity}
replaces GLUE inputs with AAVE variants, directly measuring performance
disparities relative to the standard benchmark. BST-Sarc and BST-Sent
from BESSTIE \citep{srirag2025besstiebenchmarksentimentsarcasm} evaluate
sarcasm detection and sentiment classification across British, Indian, and
Australian English, and constitute the only available coverage specific to
our three target varieties among existing English dialect benchmarks.

\subsection{Generation Evaluation}
\label{sec:gen-eval-setup}

Benchmarks measure performance under dialectal \textit{input} but do not reveal whether models produce dialectally marked \textit{output}. We, therefore, conduct a dedicated generation evaluation using \Llama{}, which shows the most consistent dialect-sensitivity gains across pipeline stages and allows more detailed qualitative examination than averaging across families would permit. We focus on the explicit thread for variety-targeted generation analysis, with the broad thread also included where alignment methods are compared directly (Section~\ref{sec:human-eval}).

\paragraph{Prompts.}
We construct a set of 25 open-ended casual prompts spanning five domains: casual conversation, opinion, food and lifestyle, hypothetical, and personal reflection, designed to elicit natural, conversational responses (Appendix~\ref{appendix:prompts}). All evaluation methods use this same prompt set.

\paragraph{Automatic Metrics.}
\textit{Variety classifier accuracy} (Appendix~\ref{app:classifier-accuracy}) is the proportion of responses correctly classified as the target variety by a DeBERTa-v3-base model fine-tuned on a deduplicated, curated split of BESSTIE \citep{srirag2025besstiebenchmarksentimentsarcasm} using an 80/10/10 stratified train/validation/test split, achieving a macro F1 of 0.766 on the held-out test set. It serves as our primary automatic proxy for holistic dialectal style.

\begin{table*}[t!]
\begin{adjustbox}{width=\linewidth, center}
\setlength{\tabcolsep}{2pt}
\renewcommand{\arraystretch}{1.1}
\begin{tabular}{l cccccc cccccc cccccc}
\toprule
& \multicolumn{6}{c}{\textbf{\Qwen{}}} & \multicolumn{6}{c}{\textbf{\Llama{}}} & \multicolumn{6}{c}{\textbf{\Gemma{}}} \\
\cmidrule(lr){2-7}\cmidrule(lr){8-13}\cmidrule(lr){14-19}
\textbf{Benchmark} & Base & CPT & SFT & DPO & GRPO & GSPO & Base & CPT & SFT & DPO & GRPO & GSPO & Base & CPT & SFT & DPO & GRPO & GSPO \\
\midrule
BBH      & 70.0 & 55.9 & 53.7 & \textbf{74.2} & 73.0 & 73.0 & 46.5 & 45.4 & 45.6 & \textbf{48.0} & 30.7 & 34.1 & \textbf{48.7} & 41.0 & 40.3 & 45.4 & 45.4 & 45.5 \\
GPQA     & 36.6 & 37.2 & 36.2 & 38.4 & 36.4 & \textbf{39.9} & \textbf{31.4} & 31.0 & 27.8 & 29.3 & 28.8 & 30.3 & 27.8 & 30.3 & 29.5 & \textbf{33.8} & 29.8 & 30.8 \\
GLUE     & \textbf{68.4} & 59.1 & 64.0 & 63.6 & 63.9 & 64.3 & \textbf{52.9} & 46.6 & 49.3 & 49.5 & 49.2 & 48.9 & \textbf{62.8} & 50.6 & 50.8 & 51.0 & 52.6 & 51.1 \\
VALUE    & 56.1 & 50.3 & 65.6 & 64.8 & \textbf{66.2} & 66.0 & 43.2 & 38.8 & 43.8 & \textbf{44.1} & 43.7 & 42.8 & \textbf{52.8} & 42.3 & 43.7 & 45.1 & 45.5 & 43.4 \\
BST-Sent & 86.1 & 89.9 & \textbf{90.2} & 90.1 & \textbf{90.2} & \textbf{90.2} & 87.1 & \textbf{87.3} & 84.7 & 85.4 & 85.3 & 85.1 & \textbf{85.2} & 82.2 & 82.5 & 78.5 & 84.4 & 82.4 \\
BST-Sarc & 37.5 & \textbf{42.8} & 41.5 & 41.3 & 40.5 & 41.3 & 38.3 & 34.9 & 38.4 & \textbf{39.0} & 38.5 & 38.9 & \textbf{34.4} & 12.5 & 25.8 & 34.2 & 24.5 & 26.0 \\
DID      & 56.4 & 61.3 & 64.9 & \textbf{65.8} & 64.6 & \textbf{65.8} & 38.9 & 40.6 & 59.3 & 50.4 & 57.1 & \textbf{61.3} & \textbf{59.4} & 35.1 & 26.4 & 13.7 & 34.6 & 27.7 \\
\bottomrule
\end{tabular}
\end{adjustbox}
\caption{Implicit adaptation thread results (\%). Base, CPT, and SFT are sequential pipeline stages; DPO, GRPO, and GSPO are alignment methods applied on top of the SFT checkpoint without dialect-specific targeting. \textbf{Bold} indicates the best result within each model family per row. BST-Sent and BST-Sarc refer to BESSTIE sentiment and sarcasm respectively.}
\label{tab:implicit-alignment}
\end{table*}

\paragraph{Human Evaluation.}
Two native or near-native speakers per dialect (six annotators total), selected on the basis of self-reported familiarity with the target variety, complete 100 blind pairwise or three-way judgements structured across four annotation tasks of 25 trials each, comparing successive pipeline stages and alignment methods (detailed in Section~\ref{sec:human-eval}). Trials are presented in blocks; model identities are not revealed. Tasks 1, 2, and 3 use A/B/Tie judgements; Task 4 is a forced three-way choice with no tie option.

\paragraph{LLM-as-a-Judge Evaluation.} We also evaluate these same four annotation tasks using \texttt{microsoft/phi-4} \citep{abdin2024phi4technicalreport}, a model outside the evaluated families to reduce architectural bias. Both conditions instruct the judge to focus on dialectal features and ignore fluency and writing quality. They differ only in the system prompt: \textit{no-features}, where Phi-4 acts as a dialect specialist without an explicit feature list; and \textit{eWAVE-features}, where the system prompt additionally provides the full eWAVE feature inventory for the target variety. Comparing conditions isolates the contribution of explicit linguistic knowledge to LLM dialectal judgements.

\section{Results}
\label{sec:results}

\begin{table*}[t]
\begin{adjustbox}{width=\linewidth, center}
\small
\renewcommand{\arraystretch}{1.35}
\setlength{\tabcolsep}{5pt}
\begin{tabular}{p{3.5cm}p{3.0cm}p{4.7cm}p{5.0cm}}
\toprule
\textbf{Config} & \textbf{Prompt} & \textbf{Broad} (\texttt{-all}) & \textbf{Explicit} (\texttt{-d}) \\
\midrule

\textbf{Model:} \Qwen{}\newline \textbf{Variety:} en-UK\newline \textbf{Adaptation:} GRPO
&
What's the best piece of advice someone's ever given you?
&
``Be kind to others, and treat them the way you would like to be treated.''\ldots{}a fundamental principle of many cultures.
&
\ldots{}This advice has helped \textbf{us} navigate through difficult situations and make decisions that align with \textbf{me} core values. \\

\midrule

\textbf{Model:} \Llama{}\newline \textbf{Variety:} en-AU\newline \textbf{Adaptation:} DPO
&
What are some small things that can brighten up a rough day?
&
\ldots{}small things, depending on what makes you feel happy and energized\ldots{}Treat yourself to a small indulgence\ldots
&
\ldots{}what makes you feel \textbf{like} happy\ldots{}Kindness to \textbf{youself}: Treat \textbf{youself} to a small indulgence\ldots \\

\midrule

\textbf{Model:} \Gemma{}\newline \textbf{Variety:} en-IN\newline \textbf{Adaptation:} GSPO
&
What's your take on people who are always late?
&
\ldots{}I understand that being late can be a frustrating experience for others\ldots{}it's not always appropriate to judge someone solely based on their punctuality\ldots
&
\ldots{}\textbf{I am understanding} that being late can be frustrating, especially where \textbf{the} punctuality is highly valued\ldots{}it's not appropriate to judge or \textbf{are criticizing} them without knowing the full contexts\ldots \\

\bottomrule
\end{tabular}
\end{adjustbox}
\caption{Qualitative comparison of broad (\texttt{-all}) and explicit (\texttt{-d}) model responses. Broad models produce dialectally neutral outputs; explicit models exhibit surface dialectal features (\textbf{bold}). en-UK: possessive \textit{me}, first-person singular \textit{us}. en-AU: \textit{like} as discourse particle, \textit{youself}. en-IN: stative progressive (\textit{am understanding}), article overextension (\textit{the} punctuality), progressive overextension (\textit{are criticizing}). Examples are selected to illustrate clear surface dialectal features and are not 
claimed to be representative of all outputs at each pipeline stage.}
\label{tab:qualitative}
\end{table*}

\subsection{Implicit and Explicit Adaptation}
\label{sec:implicit-adaptation}
In the implicit thread (Table~\ref{tab:implicit-alignment}), CPT disrupts general benchmark performance, SFT accounts for the majority of subsequent recovery, and alignment produces inconsistent shifts with no method reliably improving over SFT. Qwen is largely robust to CPT and gains on several dialect-sensitive metrics, while Gemma is most affected and fails to fully recover DID after SFT. Alignment effects are model- and benchmark-specific rather than systematic: Qwen BBH under DPO not only recovers from CPT and SFT degradation but exceeds the base model, suggesting that dialectal alignment can in some conditions enhance general reasoning, though the basis for this gain is not established and it is not replicated across families. Conversely, Llama BBH regresses notably under GRPO and GSPO, and Gemma DID collapses under DPO. Llama DID under GSPO is a notable exception where alignment adds meaningfully over SFT on a dialect task.

Table~\ref{tab:explicit-alignment} shows that in the explicit thread, \SFTd{} generally sets a strong performance baseline and alignment methods produce mostly small, inconsistent shifts. General capability benchmarks remain stable across all methods, suggesting alignment does not degrade general capability in the explicit thread. DPO shows the widest variance across varieties, with a consistent tendency to hurt on VALUE. Across varieties, no single method dominates and benchmark differences alone do not cleanly distinguish the alignment approaches, motivating the subsequent generation quality analysis in Section~\ref{sec:rqgap} onwards.

\subsection{Reward Optimisation vs.\ Perceived Dialect Quality}
\label{sec:metric-divergence}
\label{sec:rqgap}

We examine dialectal generation through automatic metrics and human and LLM judgements, with representative qualitative examples shown in Table~\ref{tab:qualitative}. As a training diagnostic, dialect feature density (Appendix~\ref{app:pipeline-metrics}) shows that GRPO most aggressively optimises the reward signal, achieving the highest eWAVE feature density across all three families. Yet variety classifier accuracy (Appendix~\ref{app:classifier-accuracy}), an independent signal covering all three model families, favours DPO for en-IN, with \SFTd{} achieving the overall highest en-AU classifier accuracy. This divergence indicates that reward optimisation and holistic dialectal quality are not equivalent: the method that maximises the training objective does not necessarily produce outputs that are more recognisably dialectal by an independent measure. An independent, reward-free linguistic analysis corroborates this pattern across all three families: the method that maximises the eWAVE reward (GRPO) produces the \emph{fewest} surface dialectal markers and the lowest feature diversity for Llama and Qwen (Appendix~\ref{app:ling}, Table~\ref{tab:ling-measures}; Figure~\ref{fig:reward-indep}). This is in part a property of the reward basis itself: eWAVE is a typological atlas documenting features that are attested across varieties, not a perceptual evaluation instrument. Some dialectal features overlap across multiple varieties, and others carry stronger associations with non-target varieties in general language use, meaning reward optimisation over this feature set can increase surface feature density without producing outputs that are holistically recognisable as the target variety. The en-UK classifier does not generalise to generated outputs (Appendix~\ref{app:classifier-accuracy}), so en-UK generation quality is assessed via human and LLM judgements only.

\subsection{Human Evaluation and LLM-as-a-Judge}
\label{sec:human-eval}

Table~\ref{tab:human-eval} reports judgement results from human annotators (two per locale) alongside two Phi-4 judge conditions, evaluated on \Llama{} outputs throughout.

The clearest finding across evaluators is that explicit adaptation produces output that human annotators recognise as dialectally distinct. Annotators judge \SFTd{} as more dialectal than the instruction-tuned baseline at high rates for en-IN and en-UK (84\% and 87\% respectively, ties excluded), and judge explicitly variety-targeted alignment as more dialectal than broad alignment at similarly high rates (71\% en-IN, 85\% en-UK). Both Phi-4 conditions agree unanimously on the latter comparison across all three varieties. These results indicate that the explicit pipeline produces perceptible dialectal features, and that variety targeting produces more recognisable output than broad alignment regardless of evaluator type. En-AU is the exception: annotators show no clear judgement between \SFTd{} and the baseline (48\%, ties excluded) and weakly judge broad alignment as more dialectal than variety-targeted in T2 (34\%). This divergence is consistent with Australian English features being more subtle due to its historical context~\cite{collins2012australian} and, consequently, less perceptible to annotators, and reinforces that explicit adaptation does not uniformly improve perceived dialectal quality across varieties.

Reward optimisation does not improve over supervised fine-tuning by any evaluator. When GRPO, the method achieving the highest automatic feature density, is compared directly against \SFTd{}, neither Phi-4
condition nor human annotators show a consistent judgement in either
direction across varieties. The high tie rate on Task~3 for en-IN (28 of
50 judgements excluded from the win rate) reflects the difficulty of
distinguishing GRPO\textsubscript{d} from \SFTd{} outputs for this variety,
consistent with the finding that reward optimisation does not reliably
produce outputs that are more recognisably dialectal. This is the clearest expression of the \rqg{} gap: alignment pushes the reward signal upward, but that movement does not translate into outputs that evaluators
find more dialectally convincing than the SFT baseline alone. Linking these judgements to output features, human judgement tracks dialectal marking
relative to the standard baseline but not reward density; annotators judge the more dialectally marked output as more dialectal than the instruct baseline yet judge the more natural, contracted output as more dialectal than GRPO (Appendix~\ref{app:ling}, Table~\ref{tab:bridge}).

\begin{table}[t]
\begin{adjustbox}{width=0.9\linewidth, center}
\renewcommand{\arraystretch}{1.2}
\setlength{\tabcolsep}{2pt}
\begin{tabular}{l l ccc}
\toprule
& & & \textbf{Phi-4} & \textbf{Phi-4} \\
\textbf{Task} & \textbf{Dial.} & \textbf{Human} & \textbf{(no ft.)} & \textbf{(+eWAVE)} \\
\midrule
\multirow{3}{*}{\textbf{T1}}
  & en-AU & 48\%~(8t)   & 43\%  & \textbf{81\%} \\
  & en-IN & \textbf{84\%}~(7t)   & \textbf{71\%}  & \textbf{96\%} \\
  & en-UK & \textbf{87\%}~(3t)   & 32\%  & \textbf{60\%} \\
\midrule
\multirow{3}{*}{\textbf{T2}}
  & en-AU & 34\%~(9t)   & \textbf{100\%} & \textbf{100\%} \\
  & en-IN & \textbf{71\%}~(8t)   & \textbf{100\%} & \textbf{100\%} \\
  & en-UK & \textbf{85\%}~(4t)   & \textbf{100\%} & \textbf{100\%} \\
\midrule
\multirow{3}{*}{\textbf{T3}}
  & en-AU & 46\%~(24t)  & 33\%  & 44\% \\
  & en-IN & 55\%~(28t)  & 32\%  & 24\% \\
  & en-UK & 38\%~(18t)  & 56\%  & 40\% \\
\midrule
\multirow{3}{*}{\textbf{T4}}
  & en-AU & GS~(36\%)   & GS~(40\%)  & GS~(48\%) \\
  & en-IN & D~(50\%)    & GS~(36\%)  & D~(56\%) \\
  & en-UK & D~(52\%)    & D~(44\%)   & D~(52\%) \\
\bottomrule
\end{tabular}
\end{adjustbox}
\caption{
    Generation judgement results across four tasks, three dialects, and three judge types.
    \textbf{T1}: Instruct vs \SFTd{} (\% \SFTd{} wins, tie-excluded).
    \textbf{T2}: GRPO$_\text{all}$ vs GRPO\textsubscript{d} (\% GRPO\textsubscript{d} wins, tie-excluded).
    \textbf{T3}: \SFTd{} vs GRPO\textsubscript{d} (\% GRPO\textsubscript{d} wins, tie-excluded).
    \textbf{T4}: method judged most dialectal and vote share (forced choice); D=DPO, GS=GSPO.
    \textbf{Human}: pooled across 2 annotators per dialect (50 judgements per task).
    Parenthesised \textit{t} = tied judgements excluded from win rates.
    \textbf{Bold}: $\geq$60\%.
}
\label{tab:human-eval}
\end{table}

A direct comparison of the alignment methods shows that GRPO is judged least dialectal across all three varieties and both Phi-4 conditions, despite achieving the highest feature density. Human annotators judge DPO as more dialectal for en-UK and en-IN, with Phi-4 in broad agreement; for en-AU, judgements favour GSPO under both judge conditions. Full inter-annotator agreement statistics are reported in Appendix~\ref{app:iaa}.

\Needspace*{6\baselineskip}
\section{Discussion}
\label{sec:discussion}
Our results highlight that dialectal robustness and generation are shaped by different pipeline components, and are not reliably co-indexed by any single method. CPT and SFT together determine the performance ceiling on standard and dialectal benchmarks; alignment operates largely within that ceiling, producing small and inconsistent benchmark shifts while visibly changing generation, a pattern confirmed generation-side by an independent linguistic analysis: surface dialectal marking is established at supervised fine-tuning and only redistributed by alignment, remaining near-absent through the base, CPT, and instruct checkpoints (Appendix~\ref{app:ling}, Figure~\ref{fig:stage-traj}).

The clearest positive finding is that explicit variety targeting consistently produces output that human annotators and LLM judges recognise as dialectally distinct, and that variety-targeted alignment is judged more dialectal than broad alignment regardless of evaluator type (Task~2). En-AU is the exception, with annotators showing no clear judgement at either stage.

The \rqg{} gap complicates this picture. GRPO achieves the highest eWAVE feature density across all three model families; yet on \Llama{}, where generation evaluation is conducted, it is the method judged least dialectal in direct comparison across all three varieties and both Phi-4 conditions (Task 4). When compared directly against \SFTd{} (Task 3), neither Phi-4 condition nor human annotators show a consistent judgement for GRPO in either direction, meaning reward optimisation does not add evaluator-perceived value over the supervised baseline. DPO, which consistently falls \textit{below} \SFTd{} in feature density, achieves better variety classifier accuracy for en-IN and is judged more dialectal by human annotators for en-UK and en-IN. This inversion follows from the reward basis itself (Section~\ref{sec:rqgap}): an inventory built to document attestation does not track perceptual dialectal quality, so raising feature density need not make output \textit{more recognisably dialectal}. 


Indian English proves consistently more challenging across most evaluation dimensions, likely reflecting the greater structural distance between IndE and the standard English training distribution and a denser active feature set (38 features) that creates more surface area for reward optimisation without producing recognisable output.

\Needspace*{6\baselineskip}
\section{Conclusion}
\label{sec:conclusion}
We presented DiaLLM, a dialect adaptation pipeline that continually pretrains three open-weight LLM families on 18 English varieties from the International Corpus of English, then applies two post-training paradigms with a systematic comparison of DPO, GRPO, and GSPO. Our goal was not to establish a single winning recipe, but to provide a controlled empirical account of what each pipeline component contributes and where current alignment methods succeed and fall short.

Two findings stand out. The first is a \rgg{} gap: benchmark performance is shaped primarily by CPT and SFT, while alignment shifts generation in ways that benchmarks do not capture. The second is a \rqg{} gap: on \Llama{}, the method that most aggressively optimises the dialectal reward signal is not the one judged most dialectal by LLM judges or human annotators; automatic evaluation across all three model families is consistent with this direction, though human and LLM judgement evidence is limited to \Llama{}. Together, these findings show two things benchmarks alone would not have revealed: that alignment reshapes generation independently of its effect on benchmark scores, and that the method which most successfully optimises a dialectal reward is not the one evaluators find most convincing.

The strongest positive result is that explicit variety targeting consistently produces output that human annotators and LLM judges recognise as dialectally distinct, and that variety-targeted alignment is consistently judged more dialectal than broad alignment regardless of evaluator type. This holds consistently for en-IN and en-UK; en-AU is the exception, with annotators showing no clear judgement at either stage. No single alignment method dominates across conditions; results vary by model family, variety, and evaluation dimension, and the patterns we observe should not be treated as general prescriptions. Gemma shows greater sensitivity to CPT disruption than Llama or Qwen, and en-IN proves consistently more challenging than en-AU or en-UK across all methods.

We release all model checkpoints and dialectal preference datasets to support future work. The varieties studied here remain substantially under-resourced at the alignment stage, and the design of effective training signals for authentic dialectal generation is an open problem. DiaLLM provides a useful empirical foundation for understanding the current limits of dialect adaptation and a starting point for the work these varieties still need.

\section*{Limitations}
\phantomsection\label{sec:limits}
Our evaluation has several limitations. First, VALUE was developed around AAVE and is not designed to benchmark our three target varieties directly; gains are best interpreted as evidence of broad dialectal robustness rather than variety-specific improvement. BESSTIE covers our target varieties but is limited to sentiment and sarcasm, leaving many dimensions of dialectal competence unassessed. Second, eWAVE was designed as a typological database for classifying variety inventories, not as a reward signal for generative training. Some attested features are pragmatically conditioned, tending to surface only in specific conversational registers or construction types that may not be elicited by generic prompts. Others represent population-level tendencies that individual speakers vary substantially on, making binary detection inherently noisy. Operationalising these features as a learnable reward therefore introduces a mismatch: the signal favours outputs that superficially satisfy classifier-detectable patterns over those that are authentically dialectal in ways the inventory was never designed to measure. Beyond this, our reward function and generation evaluation share the same eWAVE-derived feature inventory, so feature density gains---retained in Appendix~\ref{app:pipeline-metrics} as a training diagnostic only---reflect alignment with the training feature space rather than independently verified authentic dialect use. The BESSTIE-trained variety classifier provides results subject to register biases in the underlying corpus, as evidenced by the en-UK classifier's near-total collapse to en-AU predictions on LLM-generated outputs (Section~\ref{sec:metric-divergence}). Additionally, Phi-4 judge reliability was not validated against an independent human gold standard beyond the agreement analysis in Appendix~\ref{app:iaa}.

The ICE corpus consists primarily of formal and semi-formal texts---legal proceedings, academic writing, scripted broadcasts---and our models may not generalise to informal or social media registers. Our pipeline is also limited to English dialects; extension to other languages would require new CPT corpora, feature inventories, and evaluation resources. Deployment also requires knowing which variety to serve, at a cost that scales per variety; adaptive selection between variety-specific and standard models remains future work. Generation evaluation is conducted on \Llama{} only; whether the qualitative findings and judgement results generalise to Qwen or Gemma remains to be established. More broadly, our pipeline depends on both ICE corpus coverage and eWAVE feature attestation, restricting applicability to comparatively well-resourced English varieties even within the language.

Human evaluation used two annotators per variety (six total). Annotators were not dialect specialists and were selected on the basis of self-reported familiarity with the target variety. All judgements were conducted blind to model identity. Overall inter-annotator agreement is modest, reflecting the inherent difficulty of perceptual dialectal judgements with a small annotator pool; full statistics are reported in Appendix~\ref{app:iaa}. En-AU agreement is weakest (mean AC1 0.02), consistent with the weak judgement results for that variety, and en-AU human judgement results should therefore be treated as indicative rather than conclusive.

Finally, the implicit and explicit adaptation threads differ not only in whether dialectal perturbation is applied during SFT but also in the training data itself, with the implicit thread using UltraFeedback and the explicit thread using the dialect-specific preference dataset. Observed generation differences between threads therefore cannot be attributed solely to variety targeting, and future work isolating these factors through controlled ablation would sharpen the interpretation.

\section*{Ethical Considerations}

This work aims to reduce linguistic inequities by improving LLM performance on under-represented English varieties. However, explicit dialect modelling carries risks: Multi-VALUE transformations may introduce stereotypical or overgeneralised patterns, and dialect-sensitive models could be misused to infer demographic or regional identity from text. DiaLLM should be interpreted as a step toward dialectal inclusivity rather than a complete mitigation of dialect bias. All model checkpoints and datasets released with this work will include documentation cards describing intended use, known limitations, and the dialectal communities represented.

\bibliography{custom}

\appendix

\section{Corpus Details}
\label{appendix:corpdeets}

\begin{table}[H]
\centering
\footnotesize
\setlength{\tabcolsep}{4pt}
\renewcommand{\arraystretch}{1.1}
\begin{tabular}{lrr}
\toprule
\textbf{ICE Variety} & \textbf{Cleaned Tokens} & \textbf{Retention (\%)} \\
\midrule
Australia              & 655,336   & 100.0 \\
Canada                 & 1,439,016 & 38.0 \\
Great Britain          & 1,002,845 & 100.0 \\
Ireland                & 1,342,878 & 61.7 \\
USA                    & 637,093   & 63.5 \\
India                  & 1,570,211 & 41.5 \\
Singapore              & 1,412,390 & 49.1 \\
Malaysia               & 382,839   & 96.7 \\
Nigeria                & 1,308,130 & 100.0 \\
Philippines            & 1,486,351 & 45.3 \\
Cameroon               & 393,254   & 74.0 \\
East Africa            & 1,105,083 & 99.8 \\
Gibraltar              & 539,240   & 78.3 \\
Hong Kong              & 1,899,811 & 37.1 \\
Jamaica                & 1,396,602 & 40.6 \\
Scotland (spoken only) & 584,070   & 100.0 \\
Trinidad \& Tobago     & 1,403,912 & 60.6 \\
Uganda                 & 1,355,122 & 34.3 \\
\midrule
\textbf{Total (18 varieties)} & \textbf{19,914,183} & \textbf{52.4} \\
\bottomrule
\end{tabular}
\caption{ICE varieties used for continual pretraining, with retained token counts after preprocessing. Raw token total: 38,012,253. Retention varies due to differences in formatting, transcription quality, and completeness.}
\label{tab:ice-varieties}
\end{table}

We source data from 18 varieties in the International Corpus of English (ICE), listed in Table~\ref{tab:ice-varieties}, selecting those with sufficiently clean and accessible text (i.e., minimal markup corruption and complete transcript availability). Where required, access was obtained by contacting individual ICE teams, and only varieties for which usable data was available were included. The corpus spans a diverse set of regional English varieties, including both `Inner Circle' (e.g., Great Britain, USA) and `Outer Circle' (e.g., India, Nigeria, Singapore) contexts.

Preprocessing removes markup, incomplete transcripts, and duplicated content. Retention rates vary substantially across corpora, reflecting differences in data quality and preprocessing requirements. For \textit{Scotland}, only spoken data is included due to inconsistencies in the written portion. The final combined corpus contains approximately 20 million tokens. This corpus forms the sole source of data for continual pretraining across all models.

\vspace{0.5em}
\begin{table*}[ht]
\small
\setlength{\tabcolsep}{5pt}
\renewcommand{\arraystretch}{1.15}
\begin{tabular}{p{1.2cm} p{5.5cm} p{2.5cm} p{1cm} p{3.5cm}}
\toprule
\textbf{Dialect} & \textbf{Feature} & \textbf{Area} & \textbf{Rtg} & \textbf{Example} \\
\midrule

\multirow{4}{*}{NorthE}
  & Object pronouns as possessives (1SG) & Pronouns      & A & He's me brother \\
  & Us for singular object referent       & Pronouns      & A & Show us them boots \\
  & Was/were generalisation               & Agreement     & B & he were thirsty \\
  & Nowt/owt (nothing/anything)           & Adverbs       & B & I've got nowt \\
\midrule

\multirow{4}{*}{IndE}
  & Progressive with stative verbs        & Tense/aspect  & A & I'm liking this \\
  & Definite article for indefinite       & Noun phrase   & A & I had the toothache \\
  & Would for future                      & Modality      & A & I would eat tomorrow \\
  & No subject–aux inversion (wh-)        & Complementation & A & What you are doing? \\
\midrule

\multirow{4}{*}{AusE}
  & Progressive with stative verbs        & Tense/aspect  & B & I'm liking this \\
  & Bare adverbs (adj as adv)             & Adverbs       & B & drive slow \\
  & Discourse marker (sentence-final)     & Discourse     & B & it was good, like \\
  & Like as quotative/focuser             & Discourse     & B & she was like\ldots \\
\bottomrule
\end{tabular}
\caption{Representative eWAVE-derived features used in the dialect feature classifier and reward function. Ratings follow the eWAVE attestation scale: A (pervasive), B (common), C (rare), D (limited structural context); X = not applicable, ? = no data.}
\label{tab:feature-subset}
\end{table*}

\section{Dialect Feature Inventory}
\label{appendix:features}

Our dialect feature classifier is based on features derived from the eWAVE typological database, spanning 135 morphosyntactic and lexical phenomena across English varieties. These features form the basis of both the dialect density metric and the masked reward used in dialect-specific adaptation.

For reward masking, we use attested feature subsets for each target variety: 20 features for AusE, 38 for IndE, and 30 for NorthE. Table~\ref{tab:feature-subset} presents a representative subset of features used for dialect-specific reward masking across the three target varieties: Northern British English (NorthE), Indian English (IndE), and Australian English (AusE).

For dialect-specific reinforcement learning, reward computation is restricted to feature subsets corresponding to the target variety, based on attested feature--dialect mappings.

\vspace{0.5em}

\section{Hyperparameters and Infrastructure}
\label{appendix:hyperparams}

All CPT and post-training experiments were conducted on NVIDIA A100 GPUs
using a SLURM-managed cluster, with the \texttt{accelerate} library for
distributed execution. All runs use bfloat16 precision, gradient
checkpointing, and a fixed random seed of 1234.

Continual pretraining uses GaLore \citep{pmlr-v235-zhao24s} with AdamW
over 64{,}000 steps, with a GaLore rank of 1024, scale 0.25, and a
subspace update interval of 500 steps. The learning rate is
$1 \times 10^{-4}$ with linear warmup over 10\% of training, weight decay
of 0, gradient clipping at 1.0, and an effective batch size of 512.

For both SFT and \SFTd{}, models are trained for 3 epochs on
chosen completions only in a single-turn instruction format, with a
learning rate of $2 \times 10^{-5}$, cosine scheduling, and 5\% warmup.

All DPO runs use 5{,}000 steps with the sigmoid loss objective and
$\beta = 0.1$. Gemma uses AdamW at $5 \times 10^{-6}$; Llama and Qwen use
8-bit AdamW at $2 \times 10^{-6}$. Maximum prompt and completion lengths
are 2048/256 for Gemma and Llama, and 1024/192 for Qwen, with a 5\%
warmup ratio.

GRPO and GSPO share the same configuration, differing only in their
importance sampling level: GSPO applies it at the sequence level, GRPO at
the token level. Both use 4 generations per prompt over 5{,}000 steps,
with sampling temperature 0.7 and top\nobreakdash-$p = 0.95$. We set
$\beta = 0.02$, $\epsilon = 3 \times 10^{-4}$, and $\epsilon_{\text{high}}
= 4 \times 10^{-4}$. Reward weights are $(0.8, 0.1, 0.1)$ for the
dialect, COMET, and cosine components respectively; all components are
$z$-score normalised using an exponential moving average with decay 0.99
and clipping at ${\pm}5$. Optimiser settings and sequence lengths follow
the DPO configuration, with weight decay 0 and gradient clipping at 1.0.

We selected $\lambda$ by running full training for Llama~3.1-8B across all implicit and explicit threads with $\lambda \in \{0.50, 0.66, 0.80\}$ and inspecting the composite reward trajectories. $\lambda = 0.50$ and $\lambda = 0.66$ showed faster saturation of the meaning-preservation components relative to the dialectal reward, while $\lambda = 0.80$ produced the most stable joint trajectory across threads. We therefore fixed $\lambda = 0.80$ for all model families and alignment methods.

\vspace{0.5em}

\section{Dialect Feature Classifier Details}
\label{appendix:classifier}

The classifier is a BERT-base encoder \citep{devlin-etal-2019-bert} with a lightweight linear projection head, trained using binary cross-entropy loss. Training data comprises approximately 148,000 sentences derived from standard NLU benchmarks \citep{wang-etal-2018-glue} and transformed into dialectal variants using Multi-VALUE \citep{ziems-etal-2023-multi}; feature labels are obtained deterministically from the transformation process. On a held-out test set, the classifier achieves moderate overall performance, with reliable detection of high-frequency morphosyntactic markers and weaker coverage of rare or context-dependent features.

Prior to reward computation we apply per-feature temperature calibration to the sigmoid outputs, performing an independent grid search over $T \in [0.5, 3.0]$ for each of the 135 features on a held-out validation set to minimise Expected Calibration Error (ECE). The resulting temperature distribution is centred around unity ($\mu=0.994$, $\sigma=0.108$), with targeted corrections for specific features: sharpening underconfident predictions (e.g., $T=0.76$) and softening overconfident ones (e.g., $T=1.78$). This calibration ensures that sigmoid probabilities used as reward signals are well-scaled and comparable across features.

\section{Dialect Feature Density Across Pipeline Stages (Training Diagnostic)}
\label{app:pipeline-metrics}

Table~\ref{tab:pipeline-metrics} reports dialect feature density as a training diagnostic: it characterises how each pipeline stage and alignment method optimises the reward signal, but is not used as a primary evaluation metric (see Section~\ref{sec:gen-eval-setup}). Results cover all three model families, averaged across the 25 evaluation prompts.

\begin{table}[h]
\centering
\small
\setlength{\tabcolsep}{5pt}
\renewcommand{\arraystretch}{1.2}
\begin{tabular}{l ccccc}
\toprule
\textbf{Family} & CPT & SFT & DPO & GRPO & GSPO \\
\midrule
\multicolumn{6}{l}{\textit{Broad thread (\texttt{-all})}} \\
Gemma   & 2.474 & 2.817 & 2.912 & \textbf{2.959} & 2.949 \\
Llama   & 2.297 & 2.890 & 2.939 & \textbf{2.950} & 2.911 \\
Qwen    & 2.834 & 2.932 & 2.950 & \textbf{2.951} & 2.941 \\
\midrule
\multicolumn{6}{l}{\textit{Explicit thread (avg.\ across en-AU, en-IN, en-UK)}} \\
Gemma   & --- & 1.570 & 1.504 & \textbf{1.628} & 1.617 \\
Llama   & --- & 1.537 & 1.469 & \textbf{1.598} & 1.552 \\
Qwen    & --- & 1.553 & 1.495 & \textbf{1.590} & 1.561 \\
\bottomrule
\end{tabular}
\caption{$\log(1 + \sum \sigma(\text{logits}))$ over active eWAVE features across pipeline stages (training diagnostic only). Broad thread uses all 135 features; explicit thread uses variety-specific subsets. \textbf{Bold} indicates highest alignment method per row. CPT is not applicable to the explicit thread as it precedes dialect-specific branching.}
\label{tab:pipeline-metrics}
\end{table}

Three observations are consistent across all three families. First, every pipeline stage contributes dialectal signal: the log-sum score increases monotonically from CPT through alignment in the broad thread, confirming that each component adds measurable reward-side signal. Second, the explicit thread is lower in absolute terms than the broad thread at equivalent stages, reflecting the smaller active feature subsets; the relative ordering across alignment methods is preserved. Third, GRPO achieves the highest score in the explicit thread across all three families, while DPO consistently falls below \SFTd{}, a pattern discussed in Section~\ref{sec:metric-divergence}.

\vspace{0.5em}

\section{Variety Classifier Accuracy Across Model Families}
\label{app:classifier-accuracy}

Table~\ref{tab:classifier-accuracy} reports variety classifier accuracy (\%) for the explicit thread across all three model families, covering the four post-SFT stages. Accuracy is the proportion of the 25 evaluation prompt responses classified as the target variety. en-UK accuracy is near zero throughout all families and methods, confirming the classifier failure noted in Section~\ref{sec:metric-divergence} and Limitations; en-UK generation quality is therefore assessed via human and LLM-as-judge evaluation only. For en-AU and en-IN, DPO tends to maintain or improve classifier accuracy relative to \SFTd{}, while GRPO and GSPO are generally lower---consistent with the finding that higher feature density does not correspond to stronger variety classifier signal.

The en-UK variety classifier fails to generalise to LLM-generated outputs, with over 80\% of responses classified as en-AU regardless of alignment method. This reflects a register distribution shift rather than a training failure: the classifier was trained on sentiment and sarcasm data rather than open-ended conversational responses. Accordingly, en-UK generation quality is not assessed via classifier accuracy and relies on human and LLM judgements instead.

\begin{table}[H]
\centering
\small
\setlength{\tabcolsep}{5pt}
\renewcommand{\arraystretch}{1.15}
\begin{tabular}{ll cccc}
\toprule
\textbf{Variety} & \textbf{Model} & \textbf{\SFTd{}} & \textbf{DPO} & \textbf{GRPO} & \textbf{GSPO} \\
\midrule
\multirow{3}{*}{en-AU}
  & Llama & \textbf{88.0} & \textbf{88.0} & 84.0 & 76.0 \\
  & Qwen  & \textbf{88.0} & 76.0          & 80.0 & 76.0 \\
  & Gemma & \textbf{92.0} & 84.0          & 80.0 & 80.0 \\
\midrule
\multirow{3}{*}{en-IN}
  & Llama & 48.0 & \textbf{64.0} & 48.0          & 40.0 \\
  & Qwen  & 52.0 & \textbf{76.0} & 36.0          & 36.0 \\
  & Gemma & 56.0 & \textbf{60.0} & \textbf{60.0} & 56.0 \\
\midrule
\multirow{3}{*}{en-UK}
  & Llama & 0.0         & \textbf{4.0} & 0.0 & 0.0 \\
  & Qwen  & 0.0         & 0.0          & 0.0 & \textbf{4.0} \\
  & Gemma & \textbf{8.0} & 0.0         & 0.0 & 0.0 \\
\bottomrule
\end{tabular}
\caption{Variety classifier accuracy (\%) for the explicit thread over 25 evaluation prompts. \textbf{Bold} indicates best result per model per variety. en-UK accuracy is near zero across all families and methods, reflecting classifier failure rather than genuine dialectal signal (see Section~\ref{sec:metric-divergence}).}
\label{tab:classifier-accuracy}
\end{table}
\section{Human Annotation}
\label{appendix:humananno}

We recruited two native human annotators for each variety, with the expectation that each would spend less than 2 hours on the full set of tasks. They consented to the use of their annotation data for this research. Any annotators who were not authors of this work have been paid for their time. All annotators were male, except one female annotator for en-UK. 

\subsection{Task Instructions}
We provided detailed instructions to our annotators on screen. The annotation process was divided into four tasks with 25 prompts per task. 
\begin{quote}
Overview: You will complete 100 judgements across 4 tasks (25 each).
Each task asks you to read responses to a prompt and judge which sounds more dialectal for your assigned variety of English.
Full instructions appear at the start of each task.
\end{quote}
For Tasks 1 to 3, the instruction snapshot is quoted below.
\begin{quote}
You will see 25 pairs of responses to the same prompt.
Select whichever response sounds more dialectal for en-X English --- closer to how a speaker of that variety might naturally respond. \textit{Here X is replaced by the language variety.}
\end{quote}
For Task 4, however, these were changed to:
\begin{quote}
You will see 25 sets of three responses to the same prompt.
Select whichever response sounds most dialectal for en-X English.
There is no tie option---pick the best of the three. \textit{Here X is replaced by the language variety.}
\end{quote}

\subsection{Generation Prompts}
\label{appendix:prompts}
\noindent The following 25 prompts are used across all generation evaluation methods (automatic metrics, human evaluation, and LLM-as-judge). Prompts are manually curated to be open-ended, conversationally natural, and free of dialect-anchored geography, technical content, or sensitive topics.

\paragraph{Casual chat}
\begin{itemize}[noitemsep,topsep=2pt]
  \item What's a good comfort meal after a rough day?
  \item What do you reckon makes a good night out with mates?
  \item What's your take on people who are always late?
  \item What are some good ways to wind down at the end of a long day?
  \item What's the best piece of advice someone's ever given you?
\end{itemize}

\paragraph{Opinion}
\begin{itemize}[noitemsep,topsep=2pt]
  \item Do you think it's better to live in the city or the countryside?
  \item What do you think about people who just order takeaway every night instead of cooking?
  \item Is it worth spending a lot of money on a gym membership, or can you stay fit without one?
  \item What's the most overrated tourist destination in your opinion?
  \item Do you think social media does more harm than good?
\end{itemize}

\paragraph{Food \& Lifestyle}
\begin{itemize}[noitemsep,topsep=2pt]
  \item What's a simple meal worth recommending to someone just learning to cook?
  \item What's a good thing to eat on a lazy Sunday morning?
  \item Any tips for eating well without spending a fortune?
  \item What's a good breakfast to start the day right?
  \item What's a dish that everyone should try at least once?
\end{itemize}

\paragraph{Hypothetical}
\begin{itemize}[noitemsep,topsep=2pt]
  \item If you could only eat one meal for the rest of your life, what would it be and why?
  \item If you had a free weekend with no plans or obligations, how would you spend it?
  \item If you could live anywhere in the world, where would you pick and why?
  \item If you could only keep three apps on your phone, which would you choose?
  \item If a friend asked you to help them move house, what would you bring to make the day easier?
\end{itemize}

\paragraph{Personal Reflection}
\begin{itemize}[noitemsep,topsep=2pt]
  \item What's a hobby worth picking up, and what makes it enjoyable?
  \item What are some small things that can brighten up a rough day?
  \item What would an ideal Saturday look like from morning to night?
  \item What's a TV show or film worth recommending to everyone?
  \item What's a surprisingly useful thing to know that most people don't think about?
\end{itemize}

\section{Inter-Annotator Agreement}
\label{app:iaa}

Inter-annotator agreement (Table~\ref{tab:iaa}) is modest overall, 
reflecting the inherent difficulty of perceptual dialectal judgements 
with a small annotator pool. Agreement is highest on average on Task~3 
(AC$_1$ 0.25--0.57; observed 48--64\%), where raters consistently 
agree that the dialect-adapted output is more marked, consistent with 
the perceptibility finding in Section~\ref{sec:human-eval}. Agreement 
is near chance on Tasks~1 and~4 (AC$_1$ $\leq$ 0.42, mostly near 
zero), reflecting that the differences alignment introduces are small 
and hard to distinguish, consistent with the \rqg{} gap. 
en-UK shows the strongest overall agreement (mean AC$_1$ 0.44), 
including strong agreement on T2 (AC$_1$ 0.66, 72\% observed), the 
one task where raters showed stronger consensus; en-AU agreement is 
weakest (mean AC$_1$ 0.02), consistent with the weak judgement
results for that variety. Because several tasks are dominated by a 
single outcome, standard chance-corrected statistics are deflated by 
the prevalence paradox; we therefore report Gwet's AC$_1$ as the 
primary statistic alongside Krippendorff's $\alpha$ and Cohen's 
$\kappa$ for transparency~\cite{krippendorff2004,gwet2008,wongpakaran2013}.

\begin{table}[h]
\centering
\small
\begin{tabular}{llcccc}
\toprule
Variety & Task & AC$_1$ & Kripp.\ $\alpha$ & Cohen's $\kappa$ & Obs.\% \\
\midrule
\multirow{4}{*}{en-AU}
 & T1 & $-$0.02 & 0.00 & $-$0.02 & 32 \\
 & T2 & $-$0.03 & $-$0.18 & $-$0.17 & 28 \\
 & T3 & 0.25 & 0.34 & 0.20 & 48 \\
 & T4 & $-$0.11 & $-$0.06 & $-$0.06 & 24 \\
\midrule
\multirow{4}{*}{en-IN}
 & T1 & $-$0.04 & $-$0.14 & $-$0.15 & 28 \\
 & T2 & 0.17 & 0.08 & 0.00 & 40 \\
 & T3 & 0.54 & 0.45 & 0.21 & 64 \\
 & T4 & $-$0.25 & 0.01 & 0.00 & 12 \\
\midrule
\multirow{4}{*}{en-UK}
 & T1 & 0.42 & 0.36 & 0.36 & 60 \\
 & T2 & 0.66 & 0.43 & 0.27 & 72 \\
 & T3 & 0.57 & $-$0.19 & $-$0.08 & 64 \\
 & T4 & 0.11 & 0.24 & 0.16 & 40 \\
\bottomrule
\end{tabular}
\caption{Inter-annotator agreement per variety and task. 
T1: Instruct vs \SFTd{}; T2: GRPO$_\text{all}$ vs GRPO\textsubscript{d}; 
T3: \SFTd{} vs GRPO\textsubscript{d}; T4: DPO/GSPO/GRPO forced choice. 
en-UK T3 illustrates the prevalence paradox: AC$_1$ 0.57 
despite $\alpha$ $-$0.19 and $\kappa$ $-$0.08, owing to 
near-unanimous rater judgement in favour of \SFTd{}.}
\label{tab:iaa}
\end{table}

\section{Dialectal Generation: Linguistic Analysis}
\label{app:ling}
The reward signal and the variety classifier (Section~\ref{sec:rqgap}) both derive from
the same eWAVE feature space, so neither is an independent check on whether adapted
models actually \emph{produce} dialectal language. We therefore analyse the generation
outputs with a separate, rule-based instrument that does not reuse the reward classifier.
All measures are computed by a released script over the $1{,}375$ evaluation outputs
(three families $\times$ full pipeline $\times$ $25$ prompts $\times$ variants).

\paragraph{Detectors.}
We detect three independent signal types: (i) \emph{lexical} markers from curated
per-variety lexicons (e.g.\ en-AU \emph{arvo}, en-IN \emph{prepone}, en-UK \emph{nowt});
(ii) \emph{orthographic} markers via explicit British/American spelling pairs (a
classifier-independent en-UK/en-AU signal); and (iii) \emph{morphosyntactic} features over
spaCy POS/dependency parses---progressive-with-stative (\emph{I am understanding}),
\emph{was/were} generalisation, possessive \emph{me}, pluralised mass nouns
(\emph{informations}), bare adverbs, and invariant tags. Detectors are high-precision by
design; reported rates are therefore lower bounds. From the firings we derive \emph{density}
(features per 1k tokens), \emph{diversity} (distinct feature types), and \emph{stacking}
(features per sentence); the latter two are not captured by the density-based reward.

\begin{figure}[t]
  \centering
  \includegraphics[width=\columnwidth]{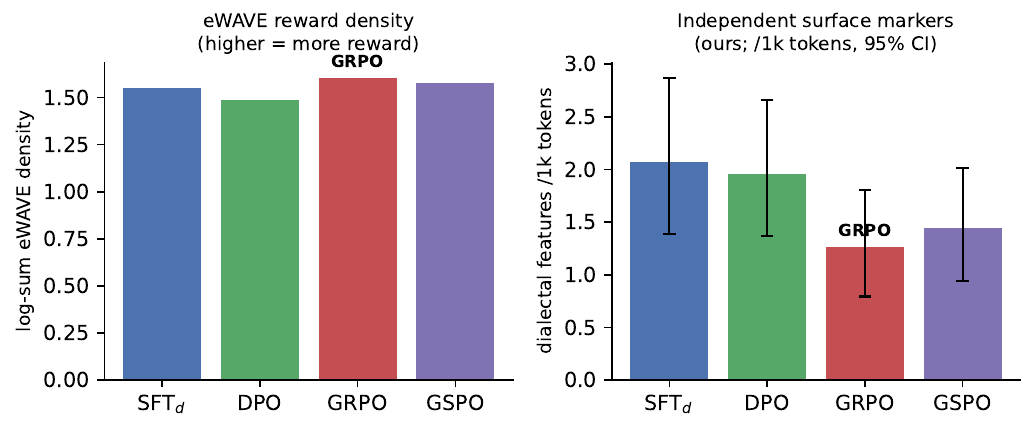}
  \caption{eWAVE \emph{reward} density (left; reproduced from Table~\ref{tab:pipeline-metrics})
  versus \emph{independent} surface-marker density (right; ours, bootstrap 95\% CIs), averaged
  over the explicit thread. GRPO is highest on the reward it optimises yet lowest on independent
  markers---the \rqg{} gap made visible.}
  \label{fig:reward-indep}
\end{figure}

\paragraph{The \rqg{} gap is corroborated by an independent measure, across families.}
Table~\ref{tab:ling-measures} and Figure~\ref{fig:reward-indep} show that, although GRPO
maximises the eWAVE \emph{reward} density (Table~\ref{tab:pipeline-metrics}), it produces the
\emph{fewest} independent surface markers for Llama and Qwen (e.g.\ Qwen density $0.87$ vs.\
\SFTd{} $2.34$ per 1k tokens) and the lowest feature diversity; the ordering is mixed only
for Gemma. These results suggest the eWAVE reward does not straightforwardly correspond to
independently measured surface dialectal richness, extending the \rqg{} gap
(Section~\ref{sec:rqgap}) beyond Llama to all three families on a non-circular measure.

\begin{figure}[t]
  \centering
  \includegraphics[width=\columnwidth]{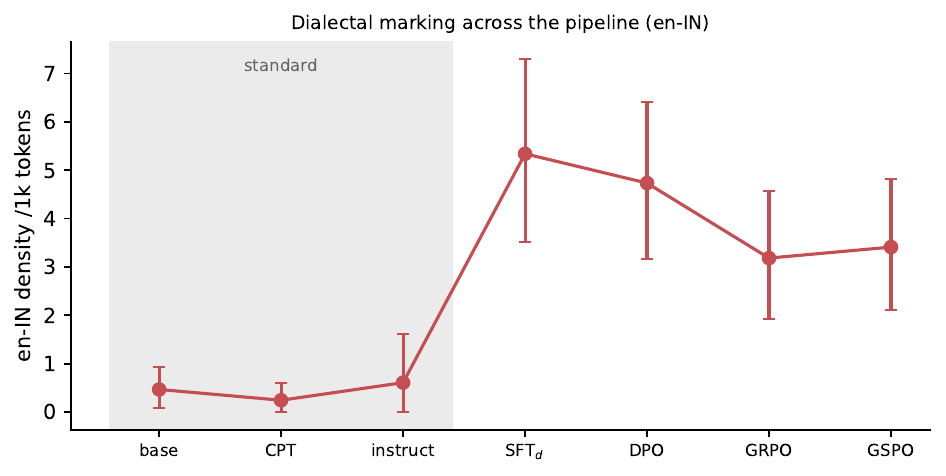}
  \caption{Dialectal-marker density across the pipeline for en-IN. Marking is near zero through
  base, CPT and instruct (standard region), rises sharply at dialectal SFT, and is thereafter
  only shifted by alignment---GRPO, the reward-maximising method, lowest (bootstrap 95\% CIs).}
  \label{fig:stage-traj}
\end{figure}

\paragraph{Alignment redistributes, rather than introduces, dialectal marking.}
Figure~\ref{fig:stage-traj} traces en-IN marker density along the pipeline: it is near zero
through the base, CPT and instruct checkpoints, rises sharply at dialectal SFT, and is
thereafter only shifted by the alignment methods, with GRPO (the method that most
optimises the reward) sitting lowest. These results suggest surface dialectal marking is
largely established by continual pretraining and supervised fine-tuning and subsequently
redistributed by alignment, consistent with the \rgg{} gap
(Section~\ref{sec:rqgap}).

\paragraph{Human judgement tracks dialect, not reward density.}
Linking the measures to the human pairwise judgements on Llama (Table~\ref{tab:bridge}),
annotators judge the higher-density output as more dialectal than the standard baseline (Task~1,
$\Delta$density $+1.38$, 95\% CI excludes $0$) but judge the \emph{more natural},
contracted output as more dialectal than the reward-maximising GRPO output (Task~3, $\Delta$contractions
$+2.87$; $\Delta$density negative, CI includes $0$). The link between dialectal density and being judged more dialectal
is therefore non-monotonic: features help against standard English, but
their reward-driven maximisation does not, consistent with the gap reported in
Section~\ref{sec:rqgap}.

\begin{figure}[t]
  \centering
  \includegraphics[width=\columnwidth]{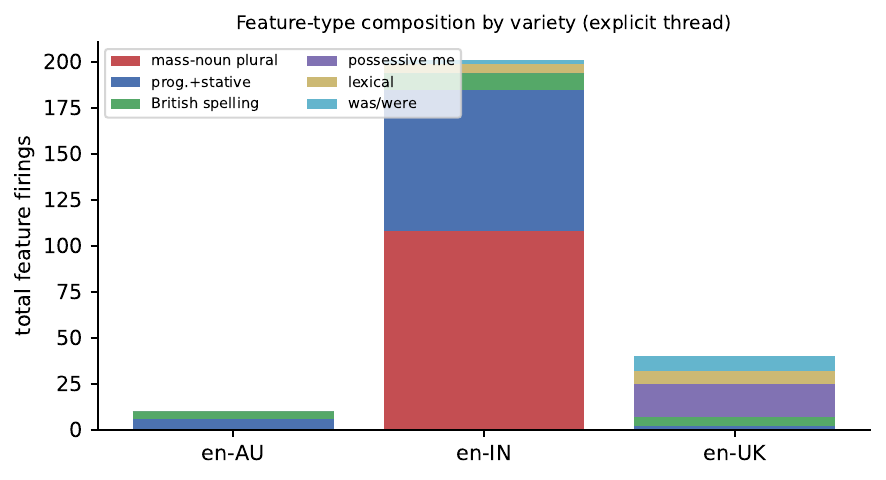}
  \caption{Feature-type composition by variety (total firings, explicit thread). en-IN marking
  is carried by mass-noun pluralisation and progressive-with-stative; en-UK by possessive
  \emph{me} and \emph{was/were} generalisation; en-AU markers are essentially absent.}
  \label{fig:feat-comp}
\end{figure}

\paragraph{Surface marking is concentrated in en-IN.}
Table~\ref{tab:variety-marking} shows that $37.7\%$ of en-IN explicit-thread outputs carry a
detected marker, against $10.0\%$ for en-UK and $3.3\%$ for en-AU; en-IN also shifts furthest
from the standard pole. The varieties also differ in \emph{which} features surface
(Figure~\ref{fig:feat-comp}): en-IN marking is carried by mass-noun pluralisation and
progressive-with-stative, whereas the few en-UK markers are predominantly possessive
\emph{me} and \emph{was/were} generalisation. The en-AU markers we detect (bare adverbs,
invariant tags) are essentially absent, consistent with the observation
(Section~\ref{sec:rqgap}) that Australian features are register-flexible and not reliably
elicited by casual prompts.

\paragraph{Caveats.}
Detectors favour precision over recall, so densities are lower bounds; en-AU/en-UK are
sparse and quantitative density claims are reliable mainly for en-IN; the human bridge is
Llama-only with modest $n$ ($48$--$85$ non-tie pairs per task), so deltas are reported with
bootstrap 95\% CIs as directional evidence. Code, detectors, and lexicons are released for
reproduction.

\begingroup\centering\small
\begin{tabular}{l rr r}
\toprule
Variety & \%\,marked & Density (/1k) & JS vs.\ std. \\
\midrule
en-AU & 3.3 & 0.16 & 0.113 \\
en-IN & 37.7 & 4.17 & 0.361 \\
en-UK & 10.0 & 0.73 & 0.142 \\
\bottomrule
\end{tabular}
\captionof{table}{Surface dialectal marking is concentrated
in en-IN. \%\,marked = share of explicit-thread outputs
with $\ge 1$ detected marker; JS = Jensen--Shannon divergence of the feature distribution from the standard
pole (base/instruct), averaged over methods/families.
en-AU markers (bare adverbs, invariant tags) are essentially absent, matching the register-flexibility note in
Section~\ref{sec:rqgap}.}
\label{tab:variety-marking}
\endgroup

\begin{table}[t]\centering\small
\begin{tabular}{ll rrr}
\toprule
Family & Method & Density & Diversity & Stacking \\
 & & (/1k) & (types) & (/sent) \\
\midrule
Llama & \SFTd{} & 1.51 & 0.19 & 0.028 \\
 & DPO & 1.77 & 0.19 & 0.033 \\
 & GRPO & 1.24 & 0.15 & 0.023 \\
 & GSPO & 1.50 & 0.19 & 0.027 \\
\addlinespace
Qwen & \SFTd{} & 2.34 & 0.25 & 0.043 \\
 & DPO & 2.32 & 0.28 & 0.043 \\
 & GRPO & 0.87 & 0.13 & 0.014 \\
 & GSPO & 1.40 & 0.19 & 0.026 \\
\addlinespace
Gemma & \SFTd{} & 2.36 & 0.21 & 0.042 \\
 & DPO & 1.79 & 0.25 & 0.033 \\
 & GRPO & 1.69 & 0.20 & 0.031 \\
 & GSPO & 1.44 & 0.19 & 0.028 \\
\addlinespace
\bottomrule
\end{tabular}
\caption{Independent surface-marker measures on the explicit thread, averaged over en-AU/en-IN/en-UK (means over $25$ prompts per cell). Density = curated dialectal features per 1k tokens; Diversity = distinct feature types; Stacking = features per sentence. In contrast to the eWAVE \emph{reward} density (Table~\ref{tab:pipeline-metrics}, where GRPO is highest), GRPO yields the \emph{lowest} independent density and diversity for Llama and Qwen---the linguistic signature of the \rqg{} gap (Section~\ref{sec:rqgap}).}
\label{tab:ling-measures}
\end{table}

\begin{table}[t]\centering\small
\setlength{\tabcolsep}{4pt}
\begin{tabular}{l rrr r}
\toprule
Task & $\Delta$Density & $\Delta$Stack & $\Delta$Contr. & $n$ \\
\midrule
T1: instruct vs \SFTd{} & \textbf{+1.38} & \textbf{+0.03} & -1.16 & 85 \\
T2: broad vs targeted & +0.78 & +0.01 & -0.02 & 83 \\
T3: \SFTd{} vs GRPO\textsubscript{d} & -0.51 & -0.01 & \textbf{+2.87} & 48 \\
\bottomrule
\end{tabular}
\caption{Human dialectal-judgement bridge (Llama). Mean winner$-$loser delta
in each linguistic measure across non-tie pairwise trials; \textbf{bold}
= bootstrap 95\% CI excludes zero. Positive = the output judged more dialectal has
\emph{more} of the measure. $n$ = non-tie trials for which both responses were
available and the measure defined. T4 is excluded: a forced three-way choice
has no winner$-$loser delta.}
\label{tab:bridge}
\end{table}

\end{document}